\documentstyle[twocolumn]{article}  

\title{Mixed Initiative in Dialogue: An Investigation into Discourse
Segmentation \\(Appeared in ACL90)}
\author{
\begin{tabular}{cc}
Marilyn Walker & Steve Whittaker\\
University of
Pennsylvania\thanks{This research was partially funded by ARO grants
DAAG29-84-K-0061 and DAAL03-89-C0031PRI, DARPA grant N00014-85-K0018, and NSF
grant
MCS-82-19196 at the University of Pennsylvania, and by Hewlett
Packard, U.K.} & Hewlett Packard Laboratories \\
Computer Science Dept. & Bristol, England BS12 6QZ \\
 Philadelphia, PA 19104 & \& HP Stanford Science Center \\
lyn@linc.cis.upenn.edu & sjw@hplb.hpl.hp.com
\end{tabular}
}
\date{}

\begin{document}           

\bibliographystyle{alpha}  
\maketitle
\begin{abstract}

Conversation between two people is usually of {\sc Mixed-Initiative},
with {\sc Control} over the conversation being transferred from one
person to another.  We apply a set of rules for the transfer of
control to 4 sets of dialogues consisting of a total of 1862 turns.
The application of the control rules lets us derive domain-independent
discourse structures. The derived structures indicate that initiative
plays a role in the structuring of discourse.  In order to explore the
relationship of control and initiative to discourse processes like
centering, we analyze the distribution of four different classes of
anaphora for two data sets. This distribution indicates that some
control segments are hierarchically related to others.  The analysis
suggests that discourse participants often mutually agree to a change
of topic. We also compared initiative in Task Oriented and Advice
Giving dialogues and found that both allocation of control and the
manner in which control is transferred is radically different for the
two dialogue types. These differences can be explained in terms of
collaborative planning principles.

\end{abstract}

\section{Introduction}

Conversation between two people has a number of characteristics that have yet
to be modeled adequately in human-computer dialogue.  Conversation is {\sc
Bidirectional}; there is a two way flow of information between participants.
Information is exchanged by {\sc Mixed-Initiative}. Each participant will, on
occasion, take the conversational lead.  Conversational partners not only
respond to what others say, but feel free to volunteer information that
is not requested and sometimes ask questions of their own\cite{Nickerson76}.
As {\sc initiative} passes back and forth between the discourse participants,
we say that {\sc Control} over the conversation gets transferred from one
discourse participant to another.

Why should we, as computational linguists, be interested in factors that
contribute to the interactivity of a discourse?  There are both theoretical and
practical motivations. First, we wish to extend formal accounts of single
utterances produced by single speakers to explain multi-participant,
multi-utterance discourses\cite{Pollack86a,CP86}.  Previous studies of
the discourse structure of multi-participant dialogues have often factored out
the role of {\sc mixed-initiative}, by allocating control to one
participant\cite{Grosz77,Cohen84a}, or by assuming a passive
listener\cite{McKeown85,RCohen87}.  Since conversation is a collaborative
process\cite{CW86,SSJ74}, models of conversation can provide the basis for
extending planning theories\cite{GS90,Cohen90}.  When the situation requires
the negotiation of a collaborative plan, these theories must account for the
interacting beliefs and intentions of multiple participants.


{}From a practical perspective, there is ample evidence that limited
mixed-initiative has contributed to lack of system usability. Many
researchers have noted that the absence of mixed-initiative gives rise
to two problems with expert systems: They don't allow users to
participate in the reasoning process, or to ask the questions they
want answered\cite{PHW82,Kidd85,FrohlichLuff89}.  In addition,
question answering systems often fail to take account of the system's
role as a conversational partner.  For example, fragmentary utterances
may be interpreted with respect to the previous user input, but what
users say is often in reaction to the system's previous
response\cite{CPA82,Sidner83b}.

In this paper we focus on interactive discourse.  We model
mixed-initiative using an utterance type classification and a set of
rules for transfer of control between discourse participants that were
proposed by Whittaker and Stenton\cite{WS88}. We evaluate the
generality of this analysis by applying the control rules to 4 sets of
dialogues, including both advisory dialogues (ADs) and task-oriented
dialogues (TODs).  We analysed both financial and support ADs. The
financial ADs are from the radio talk show ``Harry Gross: Speaking of
Your Money''\footnote{10 randomly selected dialogues (474 turns) from
a corpus that was collected and transcribed by Martha Pollack and
Julia Hirschberg\cite{HL87,PHW82}.}. The support ADs resulted from a
client phoning an expert to help them diagnose and repair various
software faults\footnote{4 dialogues (450 turns) from tapes made at
one of Hewlett-Packard's customer response centers.  See \cite{WS88}.}.
The TODs are about the construction of a plastic water pump in both
telephone and keyboard modality\footnote{5 keyboard (224 turns) and 5
telephone dialogues (714 turns), which were collected in an experiment
by Phil Cohen to explore the relationship between modality,
interactivity and use of referring expressions\cite{Cohen84a}.}.

The application of the control rules to these dialogues lets us derive
domain-independent discourse segments with each segment being
controlled by one or other discourse participant. We propose that
control segments correspond to different subgoals in the evolving
discourse plan. In addition, we argue that various linguistic devices
are necessary for conversational participants to coordinate their
contributions to the dialogue and agree on their mutual beliefs with
respect to a evolving plan, for example, to agree that a particular
subgoal has been achieved. A final phenomenon concerns shifts of
control and the devices used to achieve this.  Control shifts occur
because it is unusual for a single participant to be responsible for
coordinating the achievement of the whole discourse plan.  When a
different participant assumes control of a discourse subgoal then a
control shift occurs and the participants must have mechanisms for
achieving this.  The control framework distinguishes instances in
which a control shift is negotiated by the participants and instances
where one participant seizes control.

This paper has two objectives:
\nopagebreak
\begin{itemize}
\item To explore the phenomenon of control in relation to {\sc attentional
state} \cite{GS86,GJW86,Sidner79}\footnote{The theory of centering, which is
part of
attentional state, depends on discourse participants' recognizing the beginning
and
end of a discourse segment\cite{BFP87,Walker89b}.}.  We predict shifts of
attentional state when shifts in control are negotiated and agreed by all
participants, but not when control is seized by one participant without the
acceptance of the others. This should be reflected in different distribution of
anaphora in the two cases.

\item To test predictions about the distribution of control in
different types of dialogues.  Because the TOD's embody the
master-slave assumption\cite{GS90}, and control is allocated to the
expert, our expectation is that control should be located exclusively
with one participant in the TODs in contrast with the ADs.

\end{itemize}

\section{Rules for the Allocation and Transfer of Control}

We use the framework for the allocation and transfer of control of Whittaker
and Stenton\cite{WS88}.  The analysis is based on a classification of
utterances into 4 types\footnote{The relationship between utterance level
meaning and discourse intentions rests on a theory of joint commitment or
shared plans\cite{GS90,Cohen90}}. These are:
\begin{itemize} \item {\bf UTTERANCE TYPES}
  \begin{itemize}
   \item {\sc Assertions}: Declarative utterances used to state facts.
        {\em Yes} and {\em No} in response
 to a question were classified as assertions on the basis that they are
 supplying information.
   \item {\sc Commands}: Utterances intended to instigate action.
	Generally imperative form, but
       could be indirect such as {\em My suggestion would be that you do ....}.
   \item {\sc Questions}: Utterances which are intended to elicit information,
          including indirect forms such as {\em I was wondering whether I
should ...}.
   \item {\sc Prompts}:
        Utterances which did not express propositional content,
              such as {\em Yeah, Okay, Uh-huh ...}.
    \end{itemize}
\end{itemize}

Note that prompts are in direct contrast to the other options that a
participant has
available at any point in the discourse.  By indicating that the speaker does
not
want the floor, prompts function on a number of levels, including the
expression
of understanding or agreement\cite{Schegloff82}.

The rules for the allocation of control are based on the utterance type
classification and allow a dialogue to be divided into segments that correspond
to
which speaker is the controller of the segment.

\begin{itemize}
\item {\bf CONTROL RULES} \\ \\
\noindent \begin{tabular}{c|l}
UTTERANCE & CONTROLLER (ICP)  \\  \hline
{\sc ASSERTION}    & SPEAKER,	unless response \\
             & to a Question \\
{\sc COMMAND}      & SPEAKER  \\
{\sc QUESTION}     & SPEAKER, unless response  \\
             & to Question or Command\\
{\sc PROMPT}       & HEARER \\
\end{tabular}
\end{itemize}

The definition of controller can be seen to correspond to the intuitions behind
the term {\sc initiating conversational participant} (ICP), who is defined as
the initiator of a given discourse segment\cite{GS86}. The {\sc other
conversational participant(s)}, OCP, may speak some utterances in a segment,
but the {\sc discourse segment purpose}, must be the purpose of the ICP.  The
control rules place a segment boundary whenever the roles of the participants
(ICP or OCP) change.  For example:

\begin{footnotesize}
\noindent Abdication Example \\
\noindent E: ``And they are, in your gen you'll find that they've
          relocated into the labelled common area'' \\
\ \ \ (ASSERT - E control)\\
C: ``That's right.'' (PROMPT - E control)\\
E: ``Yeah'' (PROMPT - E abdicates control)\\
------- CONTROL SHIFT TO C --------- \\
C: ``I've got two in there. There are two of them.'' (ASSERT - C control)\\
E: ``Right'' (PROMPT - C control)\\
C: ``And there's another one which is \% RESA'' \\
\ \ \  (ASSERT - C control)\\
E: ``OK um'' (PROMPT - C control)\\
C: ``VS'' (ASSERT - C control)\\
E: ``Right'' (PROMPT - C control) \\
C: ``Mm'' (PROMPT - C abdicates control) \\
------- CONTROL SHIFT TO E  --------- \\
E: ``Right and you haven't got - I assume you haven't got
    local labelled common with those labels'' \\
\ \ \  (QUESTION - E control)\\

\end{footnotesize}

Whittaker and Stenton also performed a post-hoc analysis of the segment
boundaries that are defined by the control rules. The
boundaries fell into one of three types:

\begin{itemize}
\item {\bf CONTROL SHIFT TYPES}
\begin{itemize}
\item {\sc Abdication}: {\em Okay, go on.}
\nopagebreak
\item {\sc Repetition/Summary}: {\em That would be my recommendation
and that will ensure that you get a logically integral set of files.}
\item {\sc Interruption}: {\em It is something new though um.}
\end{itemize}
\end{itemize}

{\sc Abdications}\footnote{Our abdication category was called prompt
by \cite{WS88}.} correspond to those cases where the controller
produces a prompt as the last utterance of the segment.  The class
{\sc Repetition/Summary} corresponds to the controller producing a
{\bf redundant} utterance. The utterance is either an exact repetition
of previous propositional content, or a summary that realizes a
proposition, {\bf P}, which could have been inferred from what came
before.  Thus orderly control shifts occur when the controller
explicitly indicates that s/he wishes to relinquish control. What
unifies {\sc abdications} and {\sc repetition/summaries} is that the
controller supplies no new propositional content.  The remaining
class, {\sc interruptions}, characterize shifts occurring when the
noncontroller displays initiative by seizing control.  This class is
more general than other definitions of Interruptions. It properly
contains cross-speaker interruptions that involve topic shift, similar
to the true-interruptions of Grosz and Sidner\cite{GS86}, as well as
clarification subdialogues\cite{Sidner83b,LA90}.

This classification suggests that the transfer of control is often a
collaborative phenomenon.  Since a noncontroller(OCP), has the option
of seizing control at any juncture in discourse, it would seem that
controllers(ICPs), are in control because the noncontroller allows it.
These observations address problems raised by Grosz and
Sidner, namely how ICPs signal and OCPs recognize segment
boundaries.  The claim is that shifts of control often do not occur
until the controller indicates the end of a discourse segment by
abdicating or producing a repetition/summary.

\section{Control Segmentation and Anaphora}
\label{anaph-sec}

To determine the relationship between the derived control segments and
{\sc attentional state} we looked at the distribution of anaphora with
respect to the control segments in the ADs.  All data were analysed
statistically by $\chi^2$ and all differences cited are significant at
the 0.05 level. We looked at all anaphors (excluding first and second
person), and grouped them into 4 classes.

\begin{itemize}
\item {\bf Classes of Anaphors}
\nopagebreak
\begin{itemize}
\item {\sc 3rd Person}: {\em it, they, them, their, she, he, her, him, his}
\item {\sc One/Some}, {\em one of them, one of those, a new one, that
one, the other one, some}
\item {\sc Deictic}: Noun phrases, e.g. {\em this, that, this NP, that
NP, those NP, these NP}
\item {\sc Event}: Verb Phrases, Sentences, Segments, e.g. {\em this, that, it}
\end{itemize}
\end{itemize}

The class {\sc Deictic} refers to deictic references to material introduced
by noun phrases, whereas the class {\sc Event} refers to material introduced
clausally.

\subsection{Hierarchical Relationships}
\label{hier-sec}

The first phenomenon we noted was that the anaphora distribution indicated that
some segments are hierarchically related to others\footnote{Similar phenomena
has been noted by many researchers in discourse
including\cite{Grosz77,Hobbs79,Sidner79,PH90}.}. This was especially apparent
in
cases where one discourse participant interrupted briefly, then immediately
passed control back to the other.

\begin{footnotesize}
\noindent Interrupt/Abdicate 1

\noindent A: ... the only way I could do that was
      to take a to take a one third down and
      to take back a mortgage   (ASSERTION) \\
-------------INTERRUPT SHIFT TO B------- \\
2. B: When you talk about one third put
     a number on it                  (QUESTION) \\
3. A: uh 15 thou     (ASSERTION, but response) \\
4. B: go ahead                        (PROMPT) \\
------------ABDICATE SHIFT BACK TO A------- \\
5. A: and then I'm a mortgage back for 36 \\
\end{footnotesize}

The following example illustrates the same point.

\begin{footnotesize}
\noindent Interrupt/Abdicate 2

1. A: The maximum amount ...  will be \$400 on
      THEIR tax return.	          (ASSERTION) \\
----------INTERRUPT SHIFT TO B------- \\
2. B: 400 for the whole year?     (QUESTION) \\
3. A: yeah it'll be 20\% (ASSERTION, but response)  \\
4. B: um hm  			  (PROMPT) \\
\nopagebreak
----------ABDICATE SHIFT BACK TO A------- \\
5. A: now if indeed THEY pay the \$2000 to
      your wife.... \\
\end{footnotesize}

The control segments as defined would treat both of these cases as
composed of 3 different segments. But this ignores the fact that
utterances (1) and (5) have closely related propositional content in
the first example, and that the plural pronoun straddles the central
subsegment with the same referents being picked out by {\em they} and
{\em their} in the second example. Thus we allowed for hierarchical
segments by treating the interruptions of 2-4 as subsegments, and
utterances 1 and 5 as related parts of the parent segments. All
interruptions were treated as embeddings in this way.  However the
relationship of the segment after the interruption to the segment
before must be determined on independent grounds such as topic or
intentional structure.

\subsection{Distribution}

Once we extended the control framework to allow for the embedding of
interrupts, we coded every anaphor with respect to whether its
antecedent lay outside or within the current segment.  These are
labelled X (cross segment boundary antecedent) NX (no cross segment
boundary), in Figure \ref{HG-fig}. In addition we break these down as
to which type of control shift occurred at the previous segment
boundary.

\begin{figure}[hbt]
\begin{footnotesize}
\begin{tabular}{p{.5in}|c|c||c|c||c|c||c|c|}
& \multicolumn{2}{c}{3rd Pers}
& \multicolumn{2}{c}{One}
& \multicolumn{2}{c}{Deictic}
& \multicolumn{2}{c}{Event} \\
&X &NX  &X &NX  & X &NX  &X &NX \\
Abdication & 1 & 105 & 0 & 10 & 13 & 27 & 7 & 18 \\ \hline \\
Summary & 3 & 33 & 0 & 4 & 3 & 5 & 2 & 5 \\ \hline \\
Interrupt & 7 & 27 & 0 & 0 & 8 & 9 & 2 & 11 \\ \hline \\
TOTAL & 11 & 165 & 0 & 14 & 24 & 41 & 11 & 34 \\ \hline \\
\end{tabular}
\end{footnotesize}
\caption{Distribution of Anaphora in Finance ADs}
\label{HG-fig}
\end{figure}

We also looked at the distribution of anaphora in the Support ADs and found
similar results.
\begin{figure}[hbt]
\begin{footnotesize}
\begin{tabular}{p{.5in}|c|c||c|c||c|c||c|c|}
& \multicolumn{2}{c}{3rd Pers}
& \multicolumn{2}{c}{One}
& \multicolumn{2}{c}{Deictic}
& \multicolumn{2}{c}{Event} \\
&X &NX  &X &NX  & X &NX  &X &NX \\
Abdication & 4 & 46 & 0 & 3 & 4 & 12 & 4 & 8 \\ \hline \\
Summary & 4 & 26 & 1 & 4 & 10 & 6 & 9 & 24 \\ \hline \\
Interrupt & 8 & 40 & 0 & 4 & 5 & 5 & 5 & 10 \\ \hline \\
TOTAL & 16 & 112 & 1 & 11 & 19 & 23 & 18 & 42 \\ \hline \\
\end{tabular}
\end{footnotesize}
\caption{Distribution of Anaphora in Support ADs}
\label{supp-fig}
\end{figure}

For both dialogues, the distribution of anaphors varies according to which
type of control shift occurred at the previous segment boundary. When we look
at the different types of anaphora, we find that third person and one anaphors
cross boundaries extremely rarely, but the event anaphors and the deictic
pronouns demonstrate a different pattern.  What does this mean?

The fact that anaphora is more likely to cross segment boundaries following
interruptions than for summaries or abdications is consistent with the control
principles.  With both summaries and abdications the speaker gives an explicit
signal
that s/he wishes to relinquish control.  In contrast, interruptions are the
unprompted attempts of the listener to seize control, often having to do with
some
`problem' with the controller's utterance. Therefore, interruptions are much
more
likely to be within topic.

But why should deixis and event anaphors behave differently from the other
anaphors?  Deixis serves to pick out objects that cannot be selected by the use
of standard anaphora, i.e.  we should expect the referents for deixis to be
outside immediate focus and hence more likely to be outside the current
segment\cite{Webber86}.  The picture is more complex for event anaphora, which
seems to serve a number of different functions in the dialogue. It is used to
talk about the past events that lead up to the current situation, {\em I did
THAT in order to move the place}.  It is also used to refer to sets of
propositions of the preceding discourse, {\em Now THAT'S a little
background\/} (cf \cite{Webber88b}). The most prevalent use, however, was to
refer to
future events or actions, {\em THAT would be the move that I would make - but
you have to
do IT the same day}.

\begin{footnotesize}
\noindent SUMMARY EXAMPLE \\
A: As far as you are concerned THAT could
      cost you more .... what's your tax
      bracket?			   (QUESTION) \\
B: Well I'm on pension Harry and my wife
      hasn't  worked at all and ..(ASSERT/RESP) \\
A: No reason at all why you can't do THAT.
                                   (ASSERTION) \\
----------SUMMARY SHIFT to B------- \\
B: See my comment was if we should throw
      even the \$2000 into an IRA or something
      for her. 			   (ASSERTION) \\
----------REPETITION SHIFT to A------- \\
A: You could do THAT too. (ASSERTION) \\
\end{footnotesize}

Since the task in the ADs is to develop a plan, speakers use event anaphora as
concise references to the plans they have just negotiated and to discuss the
status and quality of plans that have been suggested.  Thus the frequent
cross-speaker references to future events and actions correspond to phases of
plan negotiation\cite{PHW82}. More importantly these references are closely
related to the control structure.  The example above illustrates the clustering
of event anaphora at segment boundaries.  One discourse participant uses an
anaphor to summarize a plan, but when the other participant evaluates this plan
there may be a control shift and any reference to the plan will necessarily
cross a control boundary.  The distribution of event anaphora bears this out,
since 23/25 references to future actions are within 2 utterances of a segment
boundary (See the example above). More significantly every instance of event
anaphora crossing a segment boundary occurs when the speaker is talking about
future events or actions.

We also looked at the TODs for instances of anaphora being used to describe a
future act in the way that we observed in the ADs.  However, over the 938 turns
in the TODs, there were only 18 instances of event anaphora, because in the
main there were few occasions when it was necessary to talk about the plan.
The financial ADs had 45 event anaphors in 474 utterances.

\section{Control and Collaborative Plans}

To explore the relationship of control to planning, we compare the TODs
with both types of ADs (financial and support). We would expect these
dialogues to differ in terms of initiative.  In the ADs, the objective is to
develop
a collaborative plan through a series of conversational exchanges.  Both
discourse participants believe that the expert has knowledge about the domain,
but
only has partial information about the situation. They also believe that the
advisee
must contribute both the problem description and also constraints as to how the
problem
can be solved.  This information must be exchanged, so that the mutual beliefs
necessary to develop the collaborative plan are established in the
conversation\cite{Joshi82}. The situation is different in the TODs.  Both
participants here believe at the outset that the expert has sufficient
information about
the situation and complete and correct knowledge about how to execute the Task.
Since
the apprentice has no need to assert information to change the expert's beliefs
or to
ask questions to verify the expert's beliefs or to issue commands, we should
not expect
the apprentice to have control.  S/he is merely present to execute the actions
indicated by the knowledgeable participant.

The differences in the beliefs and knowledge states of the participants can
be interpreted in the terms of the collaborative planning principles of
Whittaker and Stenton\cite{WS88}.  We generalize the principles of {\sc
Information Quality} and {\sc Plan Quality}, which predict when an
interrupt should occur.

\begin{itemize}

\item{\sc Information quality}: The listener must believe that the
information that the speaker has provided is true, unambiguous and relevant
to the mutual goal. This corresponds to the two rules: (A1) {\sc Truth}: If
the listener believes a fact P and believes that fact to be relevant and
either believes that the speaker believes not P or that the speaker does not
know P then interrupt; (A2){\sc Ambiguity}: If the listener believes that the
speaker's assertion is relevant but ambiguous then interrupt.

\item{\sc Plan quality}: The listener must believe that the action proposed
 by the speaker is a part of an adequate plan to achieve the mutual goal and
the action must also be comprehensible to the listener.  The two rules to
express this are: (B1){\sc Effectiveness:} If the listener believes P and
either believes that P presents an obstacle to the proposed plan or believes
that P is part of the proposed plan that has already been satisfied, then
interrupt; (B2) {\sc Ambiguity}: If the listener believes that an assertion
about the proposed plan is ambiguous, then interrupt.

\end{itemize}

These principles indirectly provide
a means to ensure mutual belief.  Since a participant must interrupt
if any condition for an interrupt holds, then lack of interruption
signals that there is no discrepancy in mutual beliefs.  If there is
such a discrepancy, the interruption is a necessary contribution to a
collaborative plan, not a distraction from the joint activity.

We compare ADs to TODs with respect to how often control is exchanged by
calculating the average number of turns between control shifts\footnote{We
excluded turns in dialogue openings and closings.}.  We also investigate
whether control is shared equally between participants and what percentage
of control shifts are represented by abdications, interrupts, and summaries
for each dialogue type. See Figure \ref{gen-fig}.

\begin{figure}
\begin{footnotesize}
\begin{tabular}{p{.65in}|c|c|c|c|}
& Finance & Support & Task-Phone & Task-Key \\
{\bf Turns/Seg} & 7.49 & 8.03  & 15.68 & 11.27  \\ \hline \\
{\bf Exp-Contr} & 60\% & 51\% & 91\% & 91\% \\ \hline \\
{\bf Abdication} & 38\% & 38\% & 45\% & 28\% \\ \hline \\
{\bf Summary} & 23\% & 27\% & 7\% & 6\% \\ \hline \\
{\bf Interrupt} & 38\% & 36\% & 48\% & 67\% \\ \hline \\
\end{tabular}

\begin{tabular}{rl}
{\bf Turns/Seg}: &Average number of turns between control shifts \\
{\bf Exp-Contr}: &\% total turns controlled by expert \\
{\bf Abdication}:& \% control shifts that are Abdications \\
{\bf  Summaries}:& \% control shifts that are Reps/Summaries  \\
{\bf  Interrupt}:& \% control shifts that are Interrupts
\end{tabular}
\end{footnotesize}
\vspace{-3.0ex}
\caption{Differences in Control for Dialogue Types}
\label{gen-fig}
\end{figure}

Three things are striking about this data.  As we predicted, the
distribution of control between expert and client is completely
different in the ADs and the TODs. The expert has control for around
90\% of utterances in the TODs whereas control is shared almost
equally in the ADs.  Secondly, contrary to our expectations, we did
find some instances of shifts in the TODs.  Thirdly, the distribution
of interruptions and summaries differs across dialogue types.  How can
the collaborative planning principles highlight the differences we
observe?

There seem to be two reasons why shifts occur in the TODs.  First, many
interruptions in the TODs result from the apprentice seizing control just
to indicate that there is a temporary problem and that plan execution
should be delayed.

\begin{footnotesize}
\noindent TASK INTERRUPT 1, A is the Instructor \\
\noindent A: It's hard to get on      (ASSERTION) \\
----------INTERRUPT SHIFT TO B------- \\
B: Not there yet - ouch yep it's there.
                              (ASSERTION) \\
A:  Okay                   (PROMPT) \\
B:  Yeah                   (PROMPT) \\
----------ABDICATE SHIFT TO A------- \\
A: All right. Now there's a little blue cap .. \\
\end{footnotesize}

Second, control was exchanged when the execution of the task started to go
awry.

\begin{footnotesize}
\noindent TASK INTERRUPT 2, A is the Instructor \\
\noindent A: And then the elbow goes over that ...
      the big end of the elbow.    (COMMAND) \\
----------INTERRUPT SHIFT TO B------- \\
B: You said that it didn't fit tight,
      but it doesn't fit tight at all,
      okay ...                    (ASSERTION) \\
A: Okay			  (PROMPT) \\
B: Let me try THIS - oops - again(ASSERTION) \\

\end{footnotesize}

The problem with the physical situation indicates to the apprentice that
the relevant beliefs are no longer shared.  The Instructor is not in
possession of critical information such as the current state of the
apprentice's pump.  This necessitates an information exchange to
resynchronize mutual beliefs, so that the rest of the plan may be
successfully executed. However, since control is explicitly allocated to
the instructor in TODs, there is no reason for that participant to believe
that the other has any contribution to make. Thus there are fewer attempts
by the instructor to coordinate activity, such as by using summaries to
synchronize mutual beliefs.  Therefore, if the apprentice needs to make a
contribution, s/he must do so via interruption, explaining why there are
many more interruptions in these dialogues.\footnote{The higher percentage
of Interruptions in the keyboard TODs in comparison with the telephone TODs
parallels Oviatt and Cohen's analysis, showing that participants exploit
the wider bandwidth of the interactive spoken channel to break tasks down
into subtasks\cite{Cohen84a,OC89a}.} In addition, the majority of
Interruptions (73\%) are initiated by apprentices, in contrast to the
ADs in which only 29\% are produced by the Clients.

Summaries are more frequent in ADs.  In the ADs both participants believe
that a plan cannot be constructed without contributions from both of them.
Abdications and summaries are devices which allow these contributions to be
coordinated and participants use these devices to explicitly set up
opportunities for one another to make a contribution, and to ensure mutual
beliefs. The increased frequency of summaries in the ADs may result from
the fact that the participants start with discrepant mutual beliefs about
the situation and that establishing and maintaining mutual beliefs is a key
part of the ADs.

\section{Discussion}
\label{disc-sec}

It has often been stated that discourse is an inherently collaborative
process and that this is manifested in certain phenomena, e.g. the use
of anaphora and cue words \cite{GS86,HL87,RCohen87} by which the
speaker makes aspects of the discourse structure explicit.  We found
shifts of attentional state when shifts in control are negotiated and
agreed by all participants, but not when control is seized by one
participant without the acceptance of the others. This was reflected
in different distribution of anaphora in the two cases.  Furthermore
we found that not all types of anaphora behaved in the same way.
Event anaphora clustered at segment boundaries when it was used to
refer to preceding segments and was more likely to cross segment
boundaries because of its function in talking about the proposed plan.
We also found that control was distributed and exchanged differently
in the ADs and TODs. These results provide support for the control
rules.

In our analysis we argued for hierarchical organization of the control
segments on the basis of specific examples of interruptions. We also
believe that there are other levels of structure in discourse that are
not captured by the control rules, e.g. control shifts do not always
correspond with task boundaries.  There can be topic shifts without
change of initiation, change of control without a topic
shift\cite{WS88}. The relationship of cue words, intonational
contour\cite{PH90} and the use of modal subordination\cite{Roberts86}
to the segments derived from the control rules is a topic for future
research.

A more controversial question concerns rhetorical relations and the extent
to which these are detected and used by listeners\cite{GS86}.  Hobbs has
applied {\sc coherence relations} to face-to-face conversation in which
mixed-initiative is displayed by participants\cite{HA85,Hobbs79}.  One
category of rhetorical relation he describes is that of {\sc Elaboration},
in which a speaker repeats the propositional content of a previous
utterance.  Hobbs has some difficulties determining the function of this
repetition, but we maintain that the function follows from the more general
principles of the control rules: speakers signal that they wish to shift
control by supplying no new propositional content.  Abdications, repetitions
and summaries all add no new information and function to signal to the
listener that the speaker has nothing further to say right now. The listener
certainly must recognize this fact.

Summaries appear to have an additional function of synchronization, by
allowing both participants to agree on what propositions are mutually
believed at that point in the discussion. Thus this work highlights aspects
of collaboration in discourse, but should be formally integrated with
research on collaborative planning\cite{GS90,Cohen90}, particularly with
respect to the relation between control shifts and the coordination of
plans.

\section{Acknowledgements}

We would like to thank Aravind Joshi for his support, comments and criticisms.
Discussions of joint action with Phil Cohen and the members of CSLI's DIA
working group have influenced the first author. We are also indebted to Susan
Brennan, Herb Clark, Julia Hirschberg, Jerry Hobbs, Libby Levison, Kathy
McKeown, Ellen Prince, Penni Sibun, Candy Sidner, Martha Pollack, Phil Stenton,
and Bonnie Webber for their insightful comments and criticisms on drafts of
this paper.

\end{document}